\documentclass[letterpaper, 10 pt, journal, twoside]{IEEEtran}

\IEEEoverridecommandlockouts

\usepackage{graphics} 
\usepackage{amsmath} 
\usepackage{amssymb}  
\usepackage[table]{xcolor}
\usepackage[ruled,linesnumbered, noend]{algorithm2e}

\usepackage{soul}
\usepackage{cite}
\usepackage{comment}
\usepackage{multirow}
\usepackage{graphicx}
\usepackage{wrapfig}
\usepackage{subfigure}
\usepackage{balance}
\usepackage{soul, color}
\usepackage{wrapfig}
\usepackage{adjustbox}

\usepackage{amsthm}

\usepackage{color, colortbl}
\definecolor{Gray}{gray}{0.9}

\newcommand{\ve}[1]{\mathbf{#1}} 
\newcommand{\tve}[1]{\tilde{\mathbf{#1}}} 

\title{TeleFMG: A Wearable Force-Myography Device for Natural Teleoperation of Multi-finger Robotic Hands}

\author{Alon Mizrahi and Avishai Sintov
\thanks{This work was supported by the Israel Science Foundation (grant No. 1565/20).}
\thanks{A. Mizrahi and A. Sintov are with the School of Mechanical Engineering, Tel-Aviv University, Israel. E-mail: {\it alonmizrahi2@mail.tau.ac.il, sintov1@tauex.tau.ac.il}.}
}

\begin{document}


\maketitle

\begin{abstract}
Teleoperation enables a user to perform dangerous tasks (e.g., work in disaster zones or in chemical plants) from a remote location. 
Nevertheless, common approaches often provide cumbersome and unnatural usage. In this letter, we propose \textit{TeleFMG}, an approach for teleoperation of a multi-finger robotic hand through natural motions of the user's hand. By using a low-cost wearable Force-Myography (FMG) device, musculoskeletal activities on the user's forearm are mapped to hand poses which, in turn, are mimicked by a robotic hand. The mapping is performed by a spatio-temporal data-based model based on the Temporal Convolutional Network. The model considers spatial positions of the sensors on the forearm along with temporal dependencies of the FMG signals. A set of experiments show the ability of a teleoperator to control a multi-finger hand through intuitive and natural finger motion. A robot is shown to successfully mimic the user's hand in object grasping and gestures. Furthermore, transfer to a new user is evaluated while showing that fine-tuning with a limited amount of new data significantly improves accuracy.
\end{abstract}


\section{Introduction}
\label{sec:introduction}

\IEEEPARstart{P}{ersonnel} in various fields are often required to balance between the need to complete dangerous and life-saving tasks and the necessity to protect themselves. These challenges exist in various hazardous domains where human operators must complete various tasks. Examples include healthcare \cite{Razu2021}, disaster zones, chemical factories, space \cite{Chen2019} and deep water \cite{Sivcev2018}. There have already been many robots working to deliver food and medicine \cite{Ozturkcan2022}, enable quarantined patients to have remote meetings with doctors and loved ones \cite{Hung2023} and inspect hazardous environments \cite{Fisher2021}. However, participation of robots, where they fully interact with the environment still lacks while they can actively perform tasks. For instance, a robot can treat a patient or handle biohazard materials. While fully autonomous robots able to manage complex tasks in hazardous environments are still some way in the future, human experts must always be in the decision process and have some control. Hence, teleoperation of robots by expert human operators are an ideal solution to put them out of harm's way.



\begin{figure}[h]
    \centering
    \includegraphics[width=\linewidth]{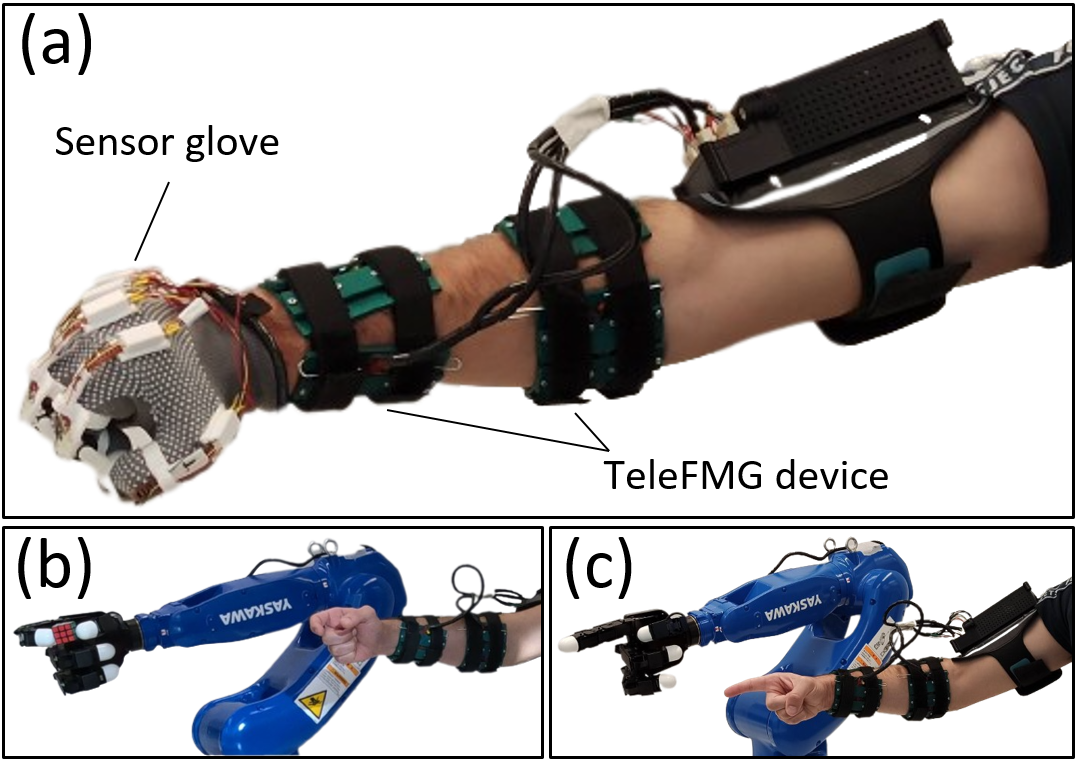}
    \caption{(a) The wearable TeleFMG device with the sensor glove are used to collect data in order to train a model for mapping musculoskeletal activities to hand poses. Then, teleoperation is demonstrated with a multi-finger robotic hand using the TeleFMG device in (b) pinch grasping of a small cube and (c) pointing gesture.}
    \label{fig:intro}
    \vspace{-0.6cm}
\end{figure}
\begin{figure*}[h]
    \centering
    \includegraphics[width=\linewidth]{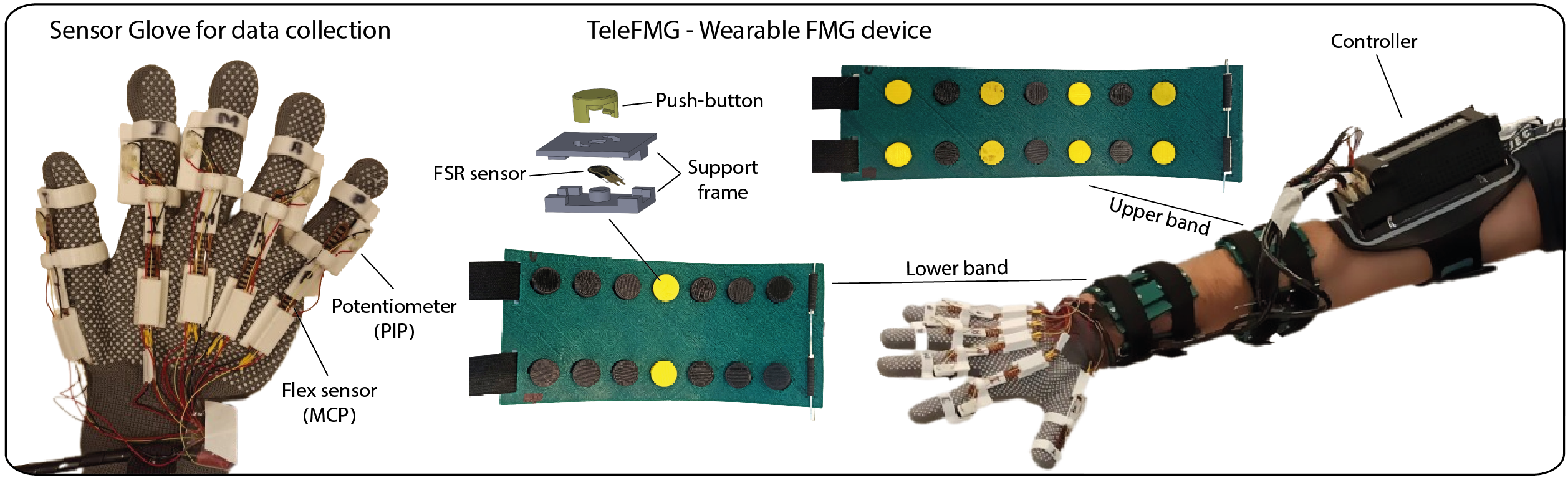}
    \caption{The TeleFMG system including two bands with 14 FSR senors each, and the sensor glove for labeling FMG signals with hand poses in the data collection phase.}
    \label{fig:System}
    \vspace{-0.5cm}
\end{figure*}

In active teleoperation, the state of the human arm and hand are mapped to motion of a robot \cite{Fani2018}. Hence, a human operator takes control over robot functions, can drive it through the environment and move a robotic arm in order to perform tasks. A common and simple approach is the use of a game-pad to move the robot \cite{Garcia2022}. Haptic teleoperation is the use of a specialized joystick or a robotic arm to control another robot along with force feedback in order to simulate the forces that the robot is experiencing \cite{Singh2020}. While haptic feedback is important, these solutions for teleoperations are usually expensive, non-intuitive, bulky and require much training \cite{Li2022}. More natural approaches directly observe the pose of the arm and hand through the use of visual perception or a sensor glove. With vision, an RGB-D camera observes the motion of the user, estimates arm and finger poses through the use of human pose estimation models and moves a robot accordingly \cite{Handa2020,Gao2023}. Relying on continuous visual perception limits the performance when visual uncertainty (e.g., poor lighting or shadows) or occlusion may occur. The use of haptic gloves is an alternative through direct measurement of hand poses \cite{Weber2016,Haptx}. While the use of sensor gloves can provide accurate finger pose estimations, they tend to be bulky and expensive while limiting the tactile sensation of the user. Hence, the user must remove them in order to perform other tasks in between teleoperation sessions. These claims are broadly addressed in a comprehensive survey on  teleoperations methods for robotic hands given in \cite{Li2022}.

A widely researched approach is to acquire and learn neurological activities through Electro-Myography (EMG) \cite{Bi2019}. EMG detects electrical signals generated by muscle tissue. For instance, an EMG was used along with an Inertial Measurement Unit (IMU) in order to teleoperate  a robotic arm and anthropomorphic five-finger hand \cite{Wolf2013}. Similarly, an EMG wrist band was proposed for control of a non-anthropomorphic robotic gripper \cite{Meeker2019}. A different work combined EMG with an haptic device to control a mobile robot \cite{Luo2020}. EMG, however, usually requires expensive and highly sizable equipment and accuracy may be compromised by electrode placement, sweat and crosstalk \cite{fujiwara2018optical}. Force-Myography (FMG), on the other hand, is an easier alternative to sense the state of a human arm \cite{amft2006}. FMG is a non-invasive technique to measure the perturbations of muscles during their contraction and relaxation. Simple force sensors on the skin of the user observe mechanical activities of muscles during motion. FMG signals were shown to be simple to acquire with a relatively high-accuracy. Consequently, FMG was used in data-based classification of hand gestures \cite{Ogris2007,Li2012,Gantenbein2023} and rehabilitation studies \cite{Yap2016}. Comparative studies have shown that FMG is less sensitive to positioning variations, does not require direct contact with the skin, and significantly outperforms EMG's accuracy \cite{Jiang2017,Belyea2019}. Previous work by the authors have shown that FMG can be used to recognize objects grasped by the human hand \cite{kahanowich2021robust} and can generalize to various new users \cite{Bamani2022}. With such information, a robot in a Human-Robot Collaboration (HRC) scenario can identify a grasped tool, infer about the intended task and act to assist.


In order to be appealing, teleoperation and interaction with a robotic system must be natural, intuitive and ergonomic. Wearable FMG offers such qualities with easy-to-use and low cost hardware. While FMG has been previously used in a wide variety of classification tasks as discussed above, it was never explored in the context of finger pose estimation for teleoperation. In this work, we explore the ability of FMG to accurately map FMG signals to the physical state of the human hand, i.e., estimate finger joint angles. We introduce the \textit{TeleFMG} system. TeleFMG is a wearable FMG device with 28 force sensors worn on the user's forearm, similar to the one proposed in \cite{kahanowich2021robust}. TeleFMG is used alongside with a data-based model in order to estimate the pose of the human fingers. The model is trained with real data labeled using a sensor glove that measures finger poses during motion. With the benchmarking of various neural-network models, we observe the required data for accurate finger pose estimation and teleoperation of a robotic system. We investigate the accuracy of a model trained with data collected from one user and to what extent can it be improved with a limited amount of new data from the new user.

In addition to accuracy evaluation, we are also interested in the ability of the model to successfully transfer tasks from the user to the robotic hand. Such tasks include hand gestures, whole hand grasping and pinching (Figure \ref{fig:intro}). With such capability, a user can remotely control a multi-finger hand. To conclude, TeleFMG offers a simple, low-cost and natural way to remotely control a robotic hand. To the best of the author's knowledge, the system is the first FMG-based technology for teleoperation of robotic hands. 

\section{System}
\label{sec:System}
In this section, we present the wearable TeleFMG device along with the sensor glove used for labeling data.

\subsection{FMG wearable device}

Previous work by the authors has proposed a low-cost wearable FMG device in the context of HRC \cite{kahanowich2021robust}. Based on the design, an advanced prototype was developed and fabricated. The device, seen in Figure \ref{fig:System}, is composed of 28 Force-Sensitive Resistors (FSR), short-tail model FSR-400 by Interlink Electronics. FSR sensors are composed of thin sheets of polymer that alter their electrical resistance in response to the amount of pressure applied to their surface. The sensors are arranged on two bands, upper and lower forearm bands, each having 14 FSR sensors equally spaced in two rows. The bands are fabricated by 3D printing with an elastic polymer (Thermoplastic elastomer). Hence, they are flexible, light-weight and allow unrestricted movement of the entire arm. To ensure compliant skin press on the FSR sensor, each sensor is covered by a push-button mechanism. The button includes an inner bulge that presses on the FSR even if the surface of the button is not parallel to the FSR. This enables adaptation on the uneven surface of the user's forearm with continuous contact. 

All FSR sensors are connected to a Teensy 4.1 micro-controller. Since the Teensy has only 18 analog channels, each two sensors of the 28 ones are connected to the same analog input in the Teensy through a voltage divider of $4.7k\Omega$ resistor. Using a transistor-based (mini MOSFET) switching system, the Teensy is able to cyclically sample different sets of sensors in a frequency of up to 40 Hz and transfer to a computing unit via cable. While not utilized in this work, a Bluetooth component was included and can handle wireless transfer of data in real-time. The total weight of the TeleFMG device including the controller is 550 grams. The cost of the device is approximated at \$150. For comparison, only a single EMG sensor can cost up to \$400.

\subsection{Sensor Glove}

The wearable device measures FMG signals from the forearm in order to map them to finger poses. A labeling system is required for recording the state of the hand. Hence, a hand labeling device based on a glove was developed. The device is composed of a cloth glove with five 2.2" SEN-10264 ROHS flex sensors and five potentiometers model PT15NH05-103A2020-S by Amphenol Piher. The flex sensors bridge the back of the hand and fingers over the knuckles, and measure the angle $\theta_{\text{MCP},i}$ of the Metacarpophalangeal (MCP) joints where $i=\{1,\ldots,5\}$ is the index of the finger. Similarly, the potentiometers measure the angle $\theta_{\text{PIP},i}$ of the Proximal Interphalangeal (PIP) joints. While the potentiometers straight-forwardly provide joint angle, the flex sensors were calibrated to map deflection to angles. In total, the sensors measure ten finger angles on the hand. In this work, we do not consider abduction and adduction motions of the fingers. The sensors are connected to the glove using 3D-printed flanges and stitches. During data collection, the labeling system is connected to a main computer along with the FMG device to allow synchronous stream of data.




\section{Method}
\label{sec:Method}
\subsection{Problem Formulation}

With the above wearable and labeling hardware, we aim to map FMG sensing to finger pose of the human hand. Let $\ve{x}\in\mathbb{R}^{28}$ be the observable state of the musculoskeletal system measured by the $28$ FSR sensors on the FMG device in contact with the forearm. Similarly, the state of the hand $\ve{q}\in\mathbb{R}^{10}$ is the set of $10$ finger joint angles where $\ve{q}=\{\theta_{\text{MCP},1},\theta_{\text{PIP},1},\ldots,\theta_{\text{MCP},5},\theta_{\text{PIP},5}\}$. The angles are zero when the fingers are fully extended. We search for a model $f$ which maps FMG signals to the pose of the hand. Since an analytical model for such map cannot be acquired, we search for a robust data-based model. With such a model, a robotic hand can mimic the motion of the human hand in real-time.


\subsection{Data Collection}
\label{sec:collection}

Training data is collected by recording FMG states through the wearable device and synchronously labeling them with hand states using the sensor glove. With a stream rate of 33 Hz, each FMG sample $\ve{x}_i$ is recorded along with its corresponding hand state $\ve{q}_i$. In this work, we wish to explore the ability of data from a single participant to provide an accurate $f$ model and to transfer to novel users. Hence, data is collected on a single participant in $n$ sessions. Before each session, the device is taken off of the forearm and re-positioned in order to collect data with positional uncertainty. In the beginning of the session and right after the FMG device was positioned on the forearm, a set of samples was taken while the participant relaxed arm muscles. The mean vector $\ve{x}_o$ of these samples is considered as the session baseline and is subtracted from any sample recorded in the session, i.e., $\tve{x}_i=\ve{x}_i-\ve{x}_o$. Such subtraction compensates for non-equal and non-uniform tightening of the device between sessions.

During a session, which included $m$ recorded samples, the participant conducted various random motions of the fingers. In order to maintain a nearly uniform data distribution, each collection session included the motion of the same set of fingers. Sessions transitioned from moving only a single finger to moving couples and so on. Since the musculoskeletal system can vary with the same finger pose but with motion of the arm and wrist, data is collected while also randomly manipulating the wrist and arm in the workspace. Sessions also included task performing such as gripping of various objects and common gestures. The resulting training data is a set of $N=mn$ labeled FMG measurements $\mathcal{P}=\{(\tve{x}_1,\ve{q}_1),\ldots,(\tve{x}_N,\ve{q}_N)\}$. A similar dataset was collected for testing trained models in independent collection sessions.





\begin{figure*}[ht]
    \centering
    \includegraphics[width=\linewidth]{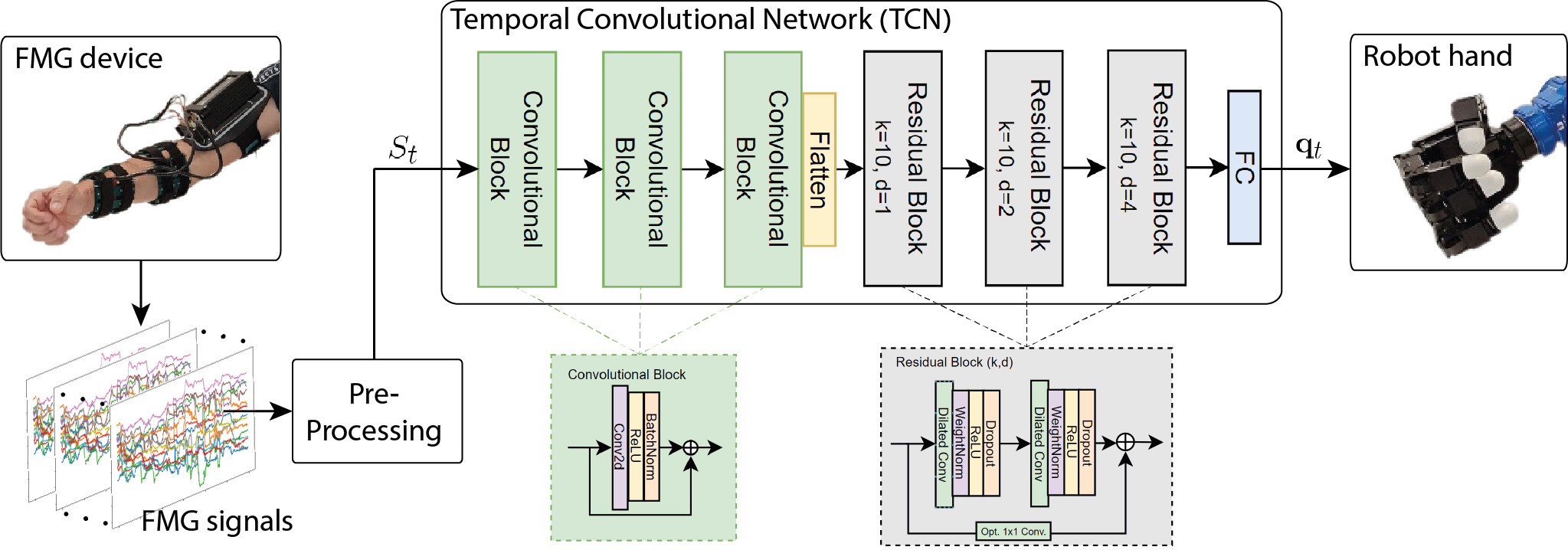}
    \caption{An illustration of the TCN model which acquires spatio-temporal FMG signals and maps them to the pose of the human hand. The pose is mimicked by a robotic hand.}
    \label{fig:TCN}
    \vspace{-0.5cm}
\end{figure*}

\subsection{Data-based Model}
\label{sec:modeling}

The presented problem requires supervised learning over dataset $\mathcal{P}$ by means of regression. One may train a Fully-Connected Neural-Network (FC-NN) to directly map a single FMG signal $\tve{x}_i$ to the corresponding pose $\ve{q}_i$ of the hand. However, it is hypothesized that temporal sequence reading of FMG signals would provide more accurate pose estimations. Let $\mathcal{S}_H\subset\mathbb{R}^{28}\times\ldots\times\mathbb{R}^{28}$ be the product space of the observable FMG space over $H$ sequential measurements. In a pre-processing step, dataset $\mathcal{P}$ is modified to include sequences of $H$ FMG signals. That is, a temporal FMG sequence of length $H$ at time $t$
\begin{equation}
    \label{eq:Sx}
    S_t=\{\tve{x}_{t-H},\ldots,\tve{x}_{t-1},\tve{x}_{t}\}\in \mathcal{S}_H
\end{equation}
is extracted from $\mathcal{P}$ and labeled with the corresponding hand pose $\ve{q}_t$. Hence, a new dataset $\mathcal{P}'=\{(S_1,\ve{q}_1),\ldots,(S_{N-H},\ve{q}_{N-H})\}$ is used to train a temporal-based model. We hypothesise that the system is govern by a map $f:\mathcal{S}_H\to\mathbb{R}^{10}$. Hence, temporal measurements from the FMG device can be mapped to the state of the hand at time $t$ through 
\begin{equation}
    \label{eq:model_eq}
    \ve{q}_t=f(S_t).
\end{equation}

Training a sequential model for \eqref{eq:model_eq} can be done using the Long Short-Term Memory (LSTM). LSTM is a class of Recurrent Neural-Networks (RNN) aimed to learn sequential data \cite{Yu2019}. LSTM is able to selectively retain or discard information from previous time steps making it well-suited for long-term dependencies. However, LSTM models can be computationally expensive and slow to train, particularly for longer sequences \cite{Orvieto2023}. Also, LSTM cannot preserve the relative positions between sensors on the FMG device. Convolutional Neural-Networks (CNN) are commonly used to learn image data and any tabular data. Hence, it is possible to formulate an array where the components are organized in the formation of the FSR sensors on the FMG device. A single channel array is then fed into the convolutional layers of the CNN. Nevertheless, such form cannot take temporal dependencies into account. In other words, it would be beneficial for a model to observe the spatial and temporal relations in the FMG signals, i.e., we require a spatio-temporal model \cite{Wang2022}. 

Each FMG signal vector $\tve{x}_{t-i}\in S_t$ is reshaped to a matrix $U_{t-i}$ of size $4\times7$. In such form, a component $u^{(a,b)}_{t-i}$ in $U_{t-i}$ denotes the FMG signal in row $a$ and column $b$ of the FMG device. Hence, matrix $U_{t-i}$ provides a spatial representation of the FMG measurement on the forearm. Consequently, a reformulated sequence 
\begin{equation}
    \label{eq:SU}
    S_t=\{U_{t-H},\ldots,U_{t-1},U_{t}\}
\end{equation}
is a spatio-temporal representation of the data. To formulate a spatio-temporal model, we use a Temporal Convolutional Network (TCN) \cite{Bai2018}. TCN is a type of NN that uses convolutional layers to process sequential data. The convolutional layers extract significant features from the data. By using dilated convolutions, the TCN is able to capture long-term dependencies in a computationally efficient manner, making it a popular and efficient choice for temporal prediction tasks.

Our proposed TCN architecture is illustrated in Figure \ref{fig:TCN}. Each matrix $U_{t-H}\in S_t$ undergoes three convolutional layers to extract meaningful features. Each convolution layer includes a ReLU activation function, batch normalization and a skip connection. After passing through the convolutional layers, the $H$ outputted matrices of size $7\times7$ are flattened and treated as a sequence of $H$ vectors. These vectors are then fed into three residual blocks which employ 1D dilated convolutions. A dilated convolutional layer allows the kernel of size $k$ to observe a wider area of the input without having to increase its size, enabling to analyze the temporal changes in the features over time. This is done by skipping $d-1$ elements between the kernel elements where $d$ is the dilation rate. The kernels of all three residual blocks have size $k=10$ while dilation rate increases to $d=1,2,4$ between layers. The proposed model was trained with $\mathcal{P}'$ while adding Gaussian noise for robustness.

\subsection{Real time work} 

Given the trained TCN model, estimated poses $\ve{q}_t$ of the human hand are to be mimicked by a multi-finger robotic hand in a teleoperation setup. Anthropomorphic robot hands usually have four- (e.g., Allegro hand) or five- (e.g., Shadow \cite{Tuffield2003} and DLR \cite{Butterfass2001} hands) fingers. In the case of a five-finger robotic hand, the finger joint angles are directly mapped to the joints of the robotic hand. When considering a four-finger robotic hand, the joint angle estimations of the little (pinky) finger are disregarded as it minimally degrades the functionality of the hand.


For each finger on the users hand, the MCP and PIP angles are estimated with FMG signals as described above. Then, they are directly mapped to the corresponding joints of the robotic hand. However, the Distal Interphalangeal (DIP) joints at the tips of the user's fingers are not measured. Yet, the human hand is known to have coupled movements termed \textit{synergies} \cite{Nestor2019}. A common representation of the synergies between the PIP and DIP of each finger is $\theta_{\text{DIP,i}}=\frac{2}{3}\theta_{\text{PIP,i}}$ \cite{ChenChen2011,Li2022}. Such ratio was used in our implementation in order to determine the DIP angles of a robotic hand based on estimation of the PIP ones on the users hand.


\section{Experiments and Results}
\label{sec:experiments}

In this section, we test and analyze the accuracy of a trained model and the ability of a user to teleoperate a robotic hand through FMG. 
Videos of data collection and experiments can be seen in the supplementary material.

\subsection{Database}

\begin{figure}[h]
    \centering
    \includegraphics[width=\linewidth]{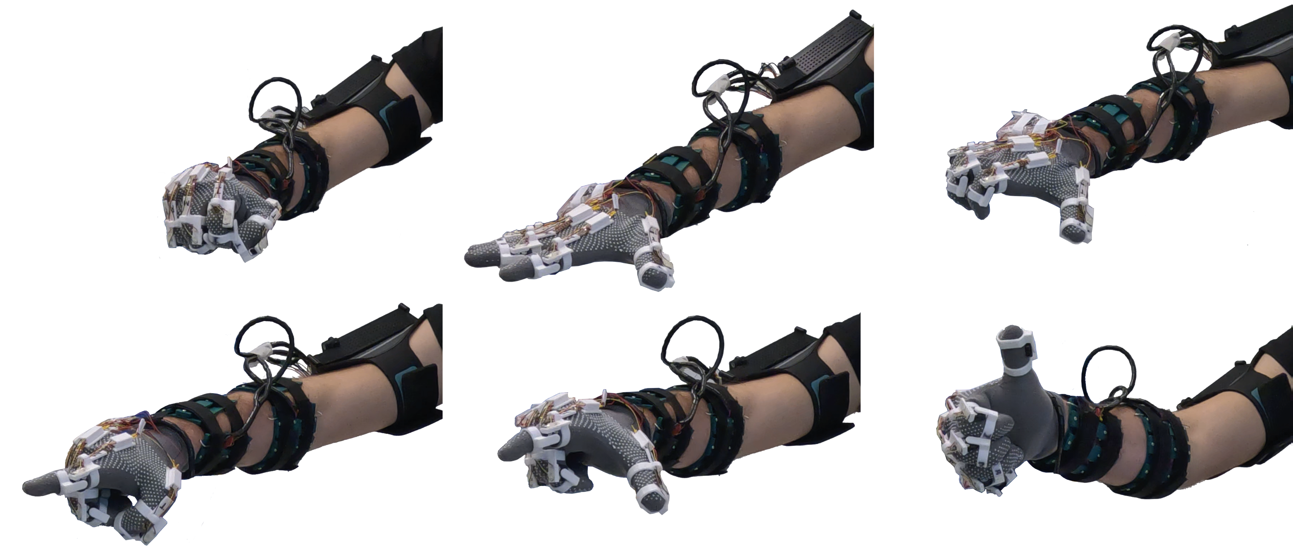}
    \caption{Snapshots of data collection using the FMG device and the labeling glove in various hand poses.}
    \label{fig:collection}
\end{figure}

Dataset $\mathcal{P}$ was collected from one participant as described in Section \ref{sec:collection} using the FMG device and labeling glove. The collection included $n=15$ sessions with $m=25,530$ recorded samples yielding a total of $N=382,950$ training samples. Such a dataset collection takes approximately four hours. The FMG device was taken off between sessions and reapplied in slightly altered locations. In addition, a test set of $105,000$ samples was recorded in $5$ separate sessions not included in the training. Snapshots from a collection session are seen in Figure \ref{fig:collection}.
\begin{table}[ht]
    \centering
    \vspace{-0.5cm}
    \caption{Estimation accuracy  for different models}
    \label{tb:accuracy}
    \begin{adjustbox}{width=0.7\linewidth}
    \begin{tabular}{lcc}
      \hline 
      Model && Angle error mean and Std. ($^{\circ}$)  \\ \hline
      5$\times$FC-NN && 14.60$~\pm~$0.80 \\
      FC-NN && 13.60$~\pm~$2.05 \\
      CNN && 15.61$~\pm~$3.37 \\
      LSTM && 13.38$~\pm~$2.51 \\
      Transformer && 13.29$~\pm~$2.27 \\

      \rowcolor{Gray}
      TCN && 9.76 $~\pm~$1.65 \\
      \hline
    \end{tabular} 
    \end{adjustbox}
    \vspace{-0.3cm}
\end{table}
\begin{table}[]
\centering
\caption{Angle mean and Std. of estimation errors ($^{\circ}$) for each finger and joint with the TCN model}
\label{tb:Fingers}
\begin{adjustbox}{width=\linewidth}
\begin{tabular}{lccccc}
    \hline
     & Thumb & Index & Middle & Ring & Little \\\hline
MCP & 12.03$\pm$2.44 & 9.92$\pm$3.24 & 9.19$\pm$2.48 & 8.65$\pm$1.86 & 8.11$\pm$2.21  \\
PIP & 12.81$\pm$3.28 & 11.6$\pm$3.31 & 8.54$\pm$2.17 & 8.31$\pm$2.58 & 8.48$\pm$2.57  \\ 
\hline
\end{tabular}
\end{adjustbox}
\end{table}

\subsection{Model evaluation}
We analyzed the performance of various deep learning models trained on dataset $\mathcal{P}$. The use of TCN as discussed in Section \ref{sec:modeling} is benchmarked with other models including: FC-NN outputting ten joint angles; five FC-NN (5$\times$FC-NN), one for each finger; LSTM; CNN; and a Transformer based on \cite{Devlin2019}. For FC-NN and 5$\times$FC-NN, the hand pose is estimated solely based on an instantaneous FMG measurement, i.e., $\ve{q}_t=f(\ve{x}_t)$. Similarly, the CNN considers instantaneous FMG measurements formulated as arrays based on sensor locations on the device. On the other hand, LSTM and Transformer can consider temporal sequences of FMG measurement as in \eqref{eq:Sx}. Hence, a sequence $S_t$ of flattened FMG signals at time $t$ is directly fed into the model to predict $\ve{q}_t$. For the TCN, the data was modified to formulate matrices as in \eqref{eq:SU} prior to training and testing. 

The hyper-parameters of all models were optimized using the Weights \& Biases package \cite{wandb} in order to provide the lowest mean errors possible. The optimal FC-NN yielded eight hidden layers with 224 neurons each and a ReLU activation function. Each network of the 5$\times$FC-NN has similar architecture to FC-NN while only having four hidden layers. The CNN includes a 4$\times$7 array inputted into two 2D convolutional layers with 3$\times$3 kernels. The output was fed into three FC layers of 324, 120 and 54 neurons with ReLU in between. For TCN, LSTM and Transformer, the optimal sequence size is $H=60$. For the LSTM, the optimization yielded a network of three layers with 54 hidden neurons each followed by three FC layers of 54, 112 and 54 neurons each with ReLU. The Transformer has four self-attention heads, five transformer blocks and two FC layers of 1,680 and 840 neurons.

Table \ref{tb:accuracy} summarizes the mean angle error over the test data and for all six models. First, momentary observation of FMG signals with FC-NN exhibits poor results. Solely re-structuring the data to learn spatial dependencies with a CNN does not provide additional accuracy. Similarly, only observing temporal dependencies in the data with an LSTM or Transformer does not provide additional accuracy improvement. By including spatial representation of the FMG data to the temporal sequencing, TCN exhibits superior results. Table \ref{tb:Fingers} presents the mean errors when individually estimating each finger joint using the TCN model. The larger errors originate from the thumb where the Carpometacarpal (CMC) joint is not modeled and provides some uncertainties. An ANOVA one-way analysis was conducted between the compared six methods in order to test statistical significance. The F-statistic resulted in $81,244$ with p-value $p<0.001$, respectively. A Tukey post-hoc test (with level of significance 0.05) revealed significant pairwise differences of TCN over other methods with a mean 4.34 and $p<0.001$, compared to an average pairwise difference of 1.11 for the other methods. Since the F-statistic is significantly high and the p-value is less than 0.05, the null hypothesis can be rejected with sufficient evidence of difference between the methods.
The results show the ability to estimate hand poses based on FMG signals and the importance of spatio-temporal inference. 


\begin{figure*}[h]
    \centering
    \includegraphics[width=\textwidth]{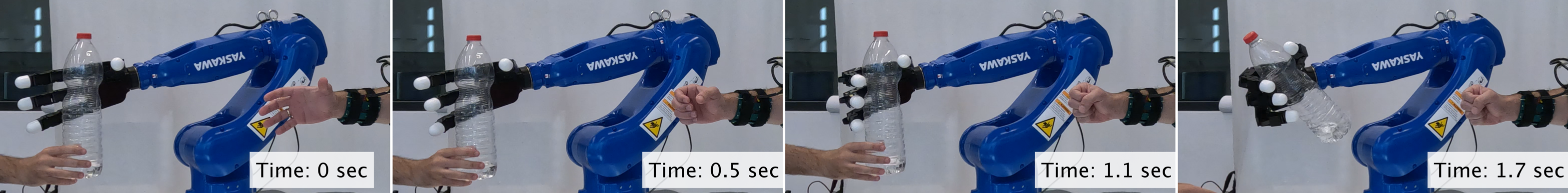}
    \caption{Snapshots of the 4-finger Allegro hand grasping a bottle using TeleFMG.}
    \label{fig:bottle}
\end{figure*}
\begin{figure*}[h]
    \centering
    \includegraphics[width=\textwidth]{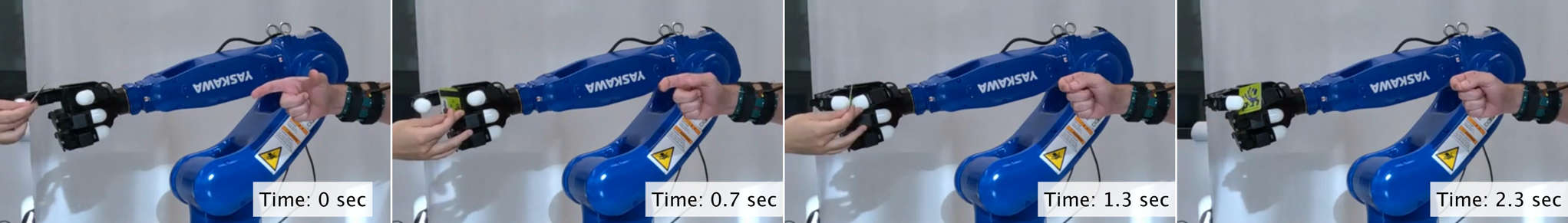}
    \caption{Snapshots of the 4-finger Allegro hand pinch grasping an ATM card using TeleFMG.}
    \label{fig:card}
\end{figure*}
\begin{figure*}[h]
    \centering
    \includegraphics[width=\textwidth]{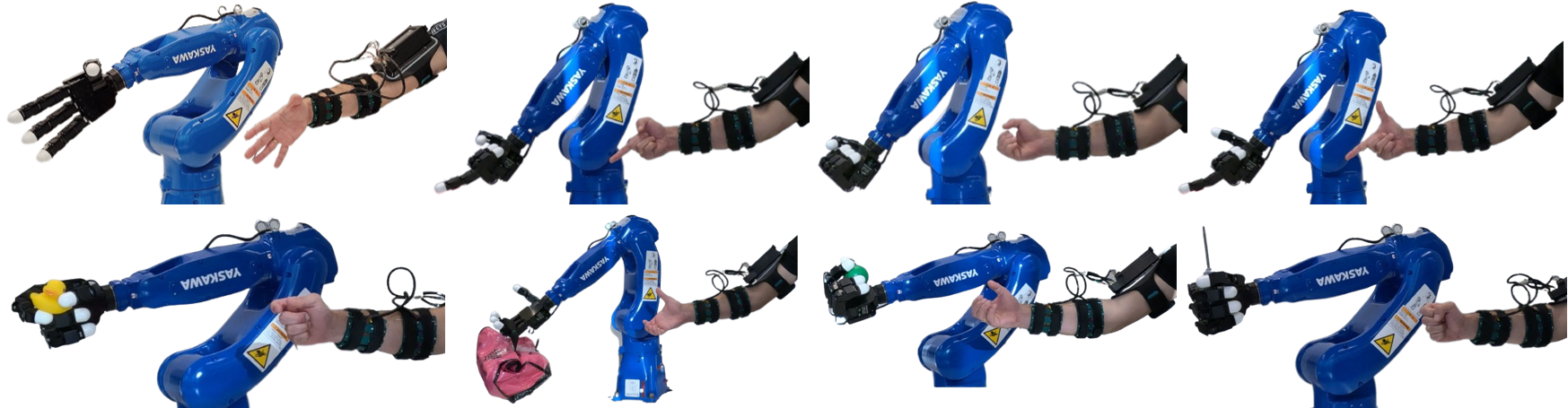}
    \caption{Snapshots of the 4-finger Allegro hand mimicking the motion of the user through TeleFMG in (top row) gestures and (bottom row) grasping various objects including (left to right) rubber duck, bag, ball and screwdriver.}
    \label{fig:snapshots}
    \vspace{-0.5cm}
\end{figure*}



\subsection{Teleoperation evaluation}

With the trained model, we wish to evaluate TeleFMG on a robotic hand. Hence, we experiment with the four-finger Allegro hand by Wonik Robotics. The Allegro is a fully-actuated hand comprised of 16 actuators, four in each finger. Three actuators on each finger control the MCP, PIP and DIP while the fourth actuate abduction and adduction motions. The latter is not used in this work and manually set to a constant value. During experiments, lagging between the motion of the user and the resultant motion of the robotic hand was noticeable. Such lag originates mainly from the low frequencies of the TeleFMG and allegro controllers. This could be solved with improved hardware. The computation frequency of the TCN model, on the other hand, is approximately $7,700$ Hz and is not a bottleneck for real-time work.

TeleFMG is first tested for teleoperation of several hand gestures including: open hand, closed hand, pointing with the index finger, thumbs-up, and a two-finger V-sign with the index and middle fingers. Then, we test the teleoperation for performing five tasks of interaction with objects including: whole hand grasp of a ball and a bottle, grasp of a thin elongated object such as a screwdriver and a brush handle, pinch grasping of small objects such as an ATM card, cube and rubber duck, and lifting bags with four fingers. The objects were handed to the robot using a human assistant. Since haptic feedback is not available in this work, the fingers stop when reached the desired angles or if their actuators reach maximum current limit. A successful task was verified visually for mimicking the gesture or safely holding the desired object. We first test the success rate of mimicking the gestures and actions of the participant whom contributed the training data.  

Table \ref{tb:teleoperation_sr} presents the success rate out of 20 attempts per task with no additional training beforehand. The total success rate is computed over all 10 tasks and 20 attempts per task. Overall, all tasks were performed with high success rate. Tasks that involve the thumb such as thumbs-up and pinch grasping failed slightly more often due to more inaccuracies of the thumb as discussed previously. This can be coped by modeling of the CMC joint in future work. Teleoperation snapshots of real-time whole hand grasping of a bottle and pinch grasping of an ATM card can be seen in Figures \ref{fig:bottle} and \ref{fig:card}, respectively. Similarly, Figure \ref{fig:snapshots} shows teleoperation demonstration of various gestures and object grasping. Based on the upcoming task, the robotic arm was manually jogged using the teach pendant to the required pose. 

\subsection{TeleFMG for new users}

The TeleFMG evaluated above was trained and tested on a single participant. We now wish to test the ability of the trained model to generalize to novel users not included in the training. Four new test participants were used for testing teleoperation while having a variety of forearm dimensions. All participants are young adults aged 25-30. The participants were instructed to perform the tasks naturally as done in non-teleoperation scenarios. Table \ref{tb:teleoperation_new_users} provides a list of anthropometric measures (i.e., forearm length (FL), lower forearm circumference (LFC) and upper forearm circumference (UFC)) for the four participants and the total success rate for performing the tasks listed in Table \ref{tb:teleoperation_sr}. Information of the train participant is also included for reference. While relatively low, the results show an ability to transfer the model to different users. User 1 has the closest anthropometric measures to the train participant and, therefore, acquired the highest success rate among the four users. On the other hand, the device did not fit well to User 3 due to a smaller forearm circumference resulting in lower success rate. These results indicate a correlation between transfer accuracy and physical measures.
\begin{table}[]
\centering
\caption{Teleoperation success rate of gestures and actions}
\label{tb:teleoperation_sr}
\begin{adjustbox}{width=0.75\linewidth}
\begin{tabular}{llc}
\hline
 & Task & Success rate (\%) \\ \hline
\multirow{5}{*}{Gestures~~~~~~~~~~}                                                                  & Open hand     & 100  \\
& Close hand    & 100  \\
& Index-point   & 95   \\
& Thumbs-up     & 80   \\
& 2-Fingers     & 90   \\ \hline
\multirow{5}{*}{Actions~~~~~~~~~~}                                                                   & Grasp ball         & 100  \\
& Grasp bottle       & 100  \\
& Grasp pole         & 100  \\
& Pinch grasp        & 85   \\
& 4-finger lift      & 95   \\ \hline
\multicolumn{1}{l}{Total} & \multicolumn{1}{l}{} & 94.5                                \\ \hline
\end{tabular}
\end{adjustbox}
\vspace{-0.2cm}
\end{table}

\begin{figure}[h]
    \centering
    \includegraphics[width=\linewidth]{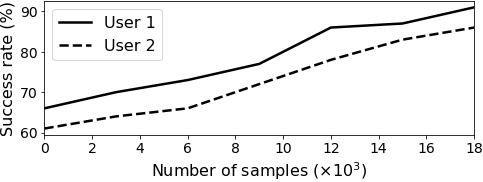}
    \vspace{-0.6cm}
    \caption{Teleoperation total task success rate in model fine-tuning for new users 1 and 2 with regards to the number of new samples from the users.}
    \label{fig:finetune}
    \vspace{-0.5cm}
\end{figure}

\begin{figure}
    \centering
    \includegraphics[width=\linewidth]{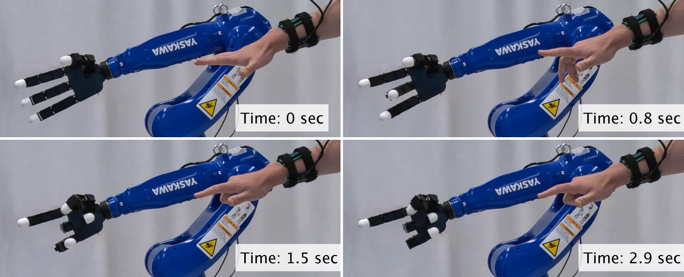}
    \caption{Demonstration of User 2 teleoperating a pointing gesture after fine-tuning the model.}
    \label{fig:user1_pointing}
\end{figure}

Inconsistencies between finger flexion magnitudes of the new user and of the train participant led to incomplete motion. Similarly, coupling between the PIP and MCP joints exhibited different proportions among users. Hence, failures to reach complete gestures occurred. In grasping, the robotic hand often failed to sufficiently close on the object yielding object drop. To improve generalization abilities, one can collect training data from various users as in \cite{Bamani2022}. Another approach, tested here, is to fine-tune the TCN model with a limited amount of training data from the new user. Given a user, we fine-tune the model with a learning rate of $10^{-5}$ and $8$ epochs with some data collected from the user using the sensor glove. Figure \ref{fig:finetune} shows success rate results after fine-tuning the model for users 1 and 2 with regards to the number of samples collected from the user. Results show that with new data of up to 5\% of the size of the original training data, the model was tuned to 91\% and 86\% for users 1 and 2, respectively. Such data collection for fine-tuning takes approximately 10 minutes. Figure \ref{fig:user1_pointing} shows a demonstration of User 1 performing a pointing gesture after the fine-tuning of the model.

\begin{table}[]
\centering
\caption{Teleoperation success rate for all users}
\label{tb:teleoperation_new_users}
\begin{adjustbox}{width=0.8\linewidth}
\begin{tabular}{ccccccc}
    \hline
    \multirow{2}{*}{User} & \multirow{2}{*}{Gender} & FL & FLC & UFC & Success rate \\
    & & (cm) & (cm) & (cm) &  (\%) \\\hline
Train & M & 28 & 17.5 & 24 & 94.5  \\
1 & M & 27 & 17   & 25 & 66  \\
2 & M & 28 & 18   & 27 & 60  \\ 
3 & F & 23 & 16.5 & 21 & 51  \\
4 & M & 30 & 19.5 & 29 & 61  \\
\hline
\end{tabular}
\end{adjustbox}
\vspace{-0.4cm}
\end{table}




\subsection{Feature Importance}

\begin{figure}
\centering
\begin{minipage}[]{0.45\linewidth}
\centering
\begin{adjustbox}{width=0.95\linewidth}
\begin{tabular}{cc|cc}
\hline
\begin{tabular}[c]{@{}c@{}}Sensor\\ index\end{tabular} & \begin{tabular}[c]{@{}c@{}}Score\\ {(}\%{)}\end{tabular} & \begin{tabular}[c]{@{}c@{}}Sensor\\ index\end{tabular} & \begin{tabular}[c]{@{}c@{}}Score\\  {(}\%{)}\end{tabular} \\ \hline
1 & 3.180  & 15 & 0.813  \\
2 & 5.816  & 16 & 2.706  \\
3 & 11.386 & 17 & 4.063  \\
4 & 1.173  & 18 & 3.123  \\
5 & 4.387  & 19 & 0.741  \\
6 & 13.013 & 20 & 2.287  \\
7 & 2.644  & 21 & 1.433  \\
8 & 2.746  & 22 & 2.016  \\
9 & 8.485  & 23 & 0.117  \\
10 & 0.254  & 24 & 0.322 \\ 
11 & 11.605 & 25 & 1.182 \\
12 & 7.841  & 26& 5.113  \\
13 & 11.583 & 27& 2.824  \\
14 & 5.083  & 28& 8.643  \\
\hline
\end{tabular}
\end{adjustbox}
\end{minipage}
\hspace{0.cm}
\begin{minipage}[]{0.5\linewidth}
\includegraphics[width=1.05\linewidth]{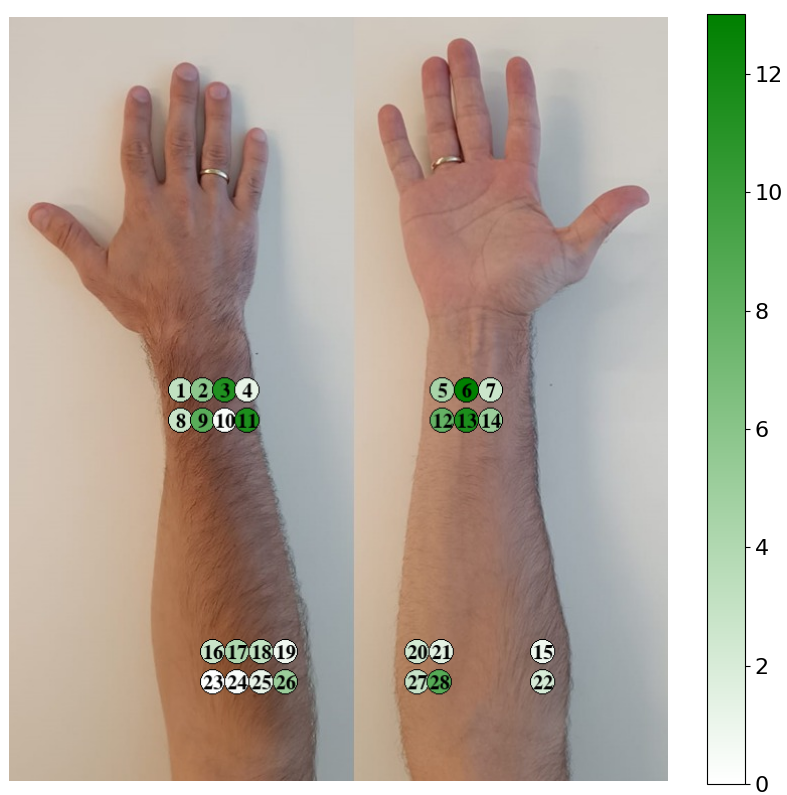}
\end{minipage}
\caption{Illustration of the sensor locations and feature importance scores of the FSR sensors on the FMG device.}
\label{fig:Sensor_performance}
\vspace{-0.5cm}
\end{figure}

We now explore the importance of the FSR sensors on joint angle prediction accuracy. Permutation feature importance is a common method to evaluate the impact of each feature in a model \cite{Yang2010}. We measure the increase in the prediction error after permuting the values of each single sensor in the test data separately. The score is the error increase resulting from the permutation of a sensor's values and is computed according to
\begin{equation}
    E_i=\frac{e_i-e}{e} \times 100\%,
\end{equation}
where $e$ is the mean error of the non-permuted model and $e_i$ is the mean error when feature $i$ is permuted. Sensor placements and sensor importance heatmap are illustrated in Figure \ref{fig:Sensor_performance}. Similar to previous results in \cite{Bamani2022}, the relative errors indicate high dependence on the lower forearm sensors and correlate with the layout of the forearm muscles. Prominent sensors 6 and 13 lay onto the flexor carpi radialis and flexor digitorum superficialis. Similarly, sensors 3 and 11 lay on the extensor carpi ulnaris. These muscles and others play an important role in operating the fingers. Overall, most sensors along the arm contribute to pose estimation accuracy. Nevertheless, one can design a new minimalistic device which focuses on the most important muscle regions that provide significant information. Furthermore, the device can assist in learning the bio-mechanics of certain human grasps.


\section{Conclusions}

In this paper, we have shown the ability of Force-Myography (FMG) on the forearm to estimate the pose of the human hand. Hence, the proposed TeleFMG enables teleoperation of robotic hands through natural motions of the human hand. In TeleFMG, a wearable FMG device is used to measure musculoskeletal activities on the forearm and map them, using a data-based model, to corresponding poses of the hand. It has been shown that a data-based model that maintains the relative spatial positions of the sensors on the forearm along with consideration of temporal dependencies provides best accuracy. Furthermore, a set of teleoperation experiments shows the ability to naturally command a multi-finger robotic hand to mimic gestures and grasping tasks. 

Future work to advance TeleFMG may consider improving accuracy by measuring additional degrees-of-freedom on the hand, particularly on the thumb. Additional work can include an IMU for telemanipulation of an entire arm in addition to the hand. Since training data was collected in various arm poses, the IMU reading could be an addition to the proposed FMG model with no requirement for modifications. Moreover, the trained model requires fine-tuning with a sensor glove in order to improve accuracy for new users. In order to cope with this limitation, a model can be trained with data collected from multiple users, as in \cite{Bamani2022}, and potentially provide zero-shot inference on new users. Wearable adaptability could also be improved in order to fit a wide range of users. Furthermore, our system cannot provide haptic feedback while important in teleoperation tasks. Hence, the addition of haptic actuators on the device can provide tactile sensation to the user upon contact of a robotic finger and applied forces. For a complete system, virtual reality goggles can be included for a sense of presence.

\bibliographystyle{IEEEtran}
\bibliography{ref}

\end{document}